# A Review of Vegetation Encroachment Detection in Power Transmission Lines using Optical Sensing Satellite Imagery

**Fathi Mahdi Elsiddig Haroun, Siti Noratiqah Mohamad Deros, Norashidah Md Din**
Institute of Energy Infrastructure Universiti Tenaga Nasional, Malaysia

**ABSTRACT**

Vegetation encroachment in power transmission lines can cause outages, which may result in severe impact on economic of power utilities companies as well as the consumer. Vegetation detection and monitoring along the power line corridor right-of-way (ROW) are implemented to protect power transmission lines from vegetation penetration. There were various methods used to monitor the vegetation penetration, however, most of them were too expensive and time consuming. Satellite images can play a major role in vegetation monitoring, because it can cover high spatial area with relatively low cost. In this paper, the current techniques used to detect the vegetation encroachment using satellite images are reviewed and categorized into four sectors; Vegetation Index based method, object-based detection method, stereo matching based and other current techniques. However, the current methods depend usually on setting manually serval threshold values and parameters which make the detection process very static. Machine Learning (ML) and deep learning (DL) algorithms can provide a very high accuracy with flexibility in the detection process. Hence, in addition to review the current technique of vegetation penetration monitoring in power transmission, the potential of using Machine Learning based algorithms are also included.

## 1. INTRODUCTION

Electrical energy is one of the most important types of energy in this century, where our daily routine depends primarily on electrical and electronics applications. Generally, electrical power distribution process goes through three main stages: the generation stage, the transmission stage, and finally, the distribution stage. The transmission stage comprises of transmission towers that are connected in a grid and plays a major role in carrying electricity over a wide geographical area. One of the challenges faced in the transmission grid are many environmental factors that may cause power interruption in transmission lines like vegetation encroachment [1]. Power outages has bad economic consequences. In 2003 for example, there was a major outage in the North America that was a result from ineffective vegetation management. Furthermore, in 2012 a survey [2] shows that 17.58% of power supply interruptions in Sarawak, Malaysia were caused by trees encroachment [3],[4]. Vegetation encroachment monitoring systems have been used to inspect power transmission lines corridor from vegetation encroachment. There were different techniques developed in order to monitor vegetation penetration in power transmission lines corridor, where the most traditional method is the manual inspection. However, this method is laborious and challenging for the inspection workers especially in difficult terrain. In the other hand, techniques like Light Detection and Ranging (LiDAR) and Synthetic Aperture Radar (SAR) have also been used to detect vegetation penetration in transmission line corridor, using Unmanned Aerial Vehicles (UAV) and airborne facilities. However, UAV and airborne drones are often too expensive in relation with the coverage area. Commercial satellite images were also used as an automated solution for vegetation penetration monitoring in wide spatial areas. Satellite observation has an advantage in vegetation encroachment monitoring, due to the lower cost and wide area inspection coverage. In addition, it can penetrate all land cover with no restrictions. Basically, there were two observation types of optical remote sensing satellite: the passive and active satellite. The passive satellite sensors observe and acquire data by using external source of radiation, such as sun radiation. On the contrary, the active sensor emits its own radiation toward the earth. There were many review and survey papers in vegetation encroachment monitoring methods for power transmission line corridor using different Optical Remote Sensing (ORS) tools like LiDAR, airborne and satellite had been published [6],[7]. In this paper, we will discuss the use of satellite images as a medium for vegetation encroachment monitoring in power transmission line right-of-way (ROW). The methods that were commonly used to analyse and manipulate satellite image to detect the vegetation penetration will also be discussed.

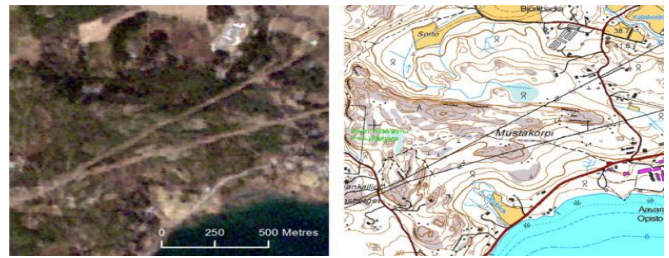

**Figure 1**: Image shows power line monitoring using satellite image[6]

## 2. SATELLITE IMAGES

In the beginning of the past decade, commercial satellite imagery has been used extensively in many applications such as climate change impact and Land Cover (LC) monitoring, where passive satellites can monitor different wavelengths of different electromagnetic spectrum (EMS) reflected from Earth's surface. Generally, the passive satellites have sensors that are capable to observe multi electromagnetic spectrum channels such as visible light and Near Infrared (NIR) which commonly known as bands. Satellite images were often categorized based on the pixel resolution scale which describes how much the Ground Sample Distance (GSD) relative to digital image pixel. The lower the GSD reflects higher resolution imageries acquired. Figure 1 shows an example for vegetation monitoring using satellite images [6] while Table 1 shows the classification of satellite images based on GSD.

Table 1 : shows satellite image classification based on GSD [6]

| Low | Medium | High | Very high |
|---|---|---|---|
| ⩾30m | ⩾ 5m < 30m | ⩾ 1m <5m | < 1m |

## 3. CURRENT DETECTION METHODS

Satellite imagery can be used to detect the vegetation penetration in power transmission line corridor. There were different methods used to detect the vegetation penetration by using satellite images includes the Vegetation Index (VI) like the Normalized Difference Vegetation (NDVI) index, Enhanced Vegetation Index (EVI) and Atmospherically Resistance Vegetation Index (ARVI).

### 3.1 Vegetation Index (VI) Based

NDVI index commonly used in satellite images to detect vegetation activity in the land. Vegetation spectral signatures are dominant in NIR where they are largely reflected. The red band, on the other hand, is heavily absorbed [8]. Figure 2 shows the optical properties of the vegetation index that are used for obtaining NDVI. NDVI is one of the major indicator for vegetation classification and land cover classification[8],[9]. NDVI can be calculated using (1) Where the NIR represent the Near-Infrared reflection from EMS, and the RED represent the red-light reflection band [10].

$$NDVI = \left(\frac{NIR - RED}{NIR + RED}\right) \quad (1)$$

Work done by [10] used satellite images to detect the Overhead Power Line (OPL) corridor in Pskov region in Russia using NDVI. The total length of the power transmission line under study was about 543 kilometers. The satellite images were from Sentinel-2. NDVI threshold values were applied on the satellite data as a vegetation penetration detection method. Using a Geographical Information System (GIS) software, and they discovered that 84% of the power line corridor needs to be managed A study on different observation techniques to monitor vegetation in power line corridor, including LiDAR, videography and satellite images were conducted by [11]. They applied the ARVI technique, which can mitigate the atmospheric effect in the observation process, to distinguish different vegetation classes in satellite images. The work used GeoMedia software, where the ARVI can be calculated as (2) where gamma ($\gamma$) represent the resistance factor, NIR is the Near Infrared, RED represent the red band and BLUE is the blue band [12] ,[13].

$$ARVI = \frac{(NIR - (RED - \gamma(BLUE - RED)))}{(NIR + (RED - \gamma(BLUE - RED)))} \quad (2)$$

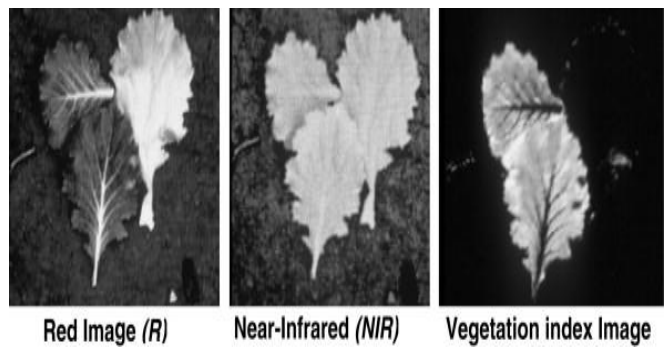

**Figure 2:** Optical properties of vegetation index [14]

### 3.2 Object Detection Based

Another method for vegetation encroachment detection inside the power line corridor is the Object Detection Based (ODB) method, where the object is often the power line corridor or the transmission tower base. [5] implemented the OBD method on Google map images, to extract and detect the power line corridor from the image. The target image that contains all the information was loaded and the image was filtered after being converted to grayscale. In order to remove any non-relevant data, the transmission tower library was created by extracting the average histogram of each tower individually. Finally, the target was extracted pixel by pixel to match it with the transmission tower library. Then the bounding box was created around the towers and a path between the transmission towers was drawn to extract the corridor from the image. Figure 3 shows the general steps for OBS.

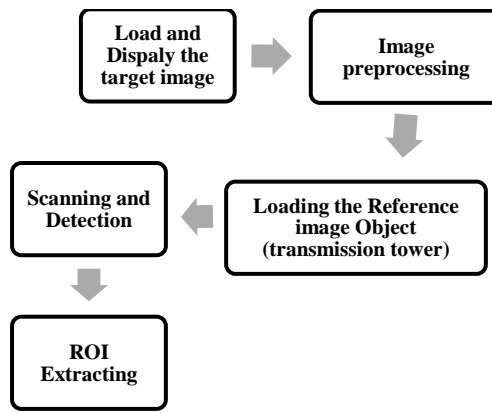

**Figure 3**: Object Based detection steps

### 3.3 Stereo Matching Detection

Stereo matching is one of the current techniques that has been used widely to detect vegetation encroachment within the power transmission line corridor using satellite images. The main idea behind stereo matching is to generate digitalized three-dimensional image that can demonstrate the difference between two images in terms of their depth or height. Figure 4 shows the stereo matching technique where there are two distinct images on the left ($I_L$) and right ($I_R$) with three location coordination (x, y, z) in the both images, and both images refers to the same point on land P (x, y, z). This technique can be used to produces a depth map by calculating the disparity between the two images, that can estimate the vegetation height when analyzed with other GIS information as the GIS data contain the transmission towers geographical locations data. Stereo matching technique was implemented in [15], where the detection of vegetation encroachment was done by identification of the tree health based on NDVI. Then the stereo matching was used to estimate the height of trees surrounding the transmission tower. The detection algorithm was programmed by using JAVA programming language. Researchers in [16] designed a vegetation detection method that depended on classifying satellite image into regions based on NDVI values. Subsequently, the vegetation height detection was done by using Digital Elevation Modelling (DEM) and Digital Surface Modelling (DSM). This study used Definiens Pro5 GIS software to perform the detection process. Before the implementation of the NDVI process, image pre-processing was carried out by performing ortho-rectification in order to adjust the satellite sensor errors. The atmospheric influence was also reduced by applying Atmospheric Correction Tool (ATCOR) [7]. Disparity maps are produced in matching stereo pairs. Several different techniques that have been used to generate disparity maps such as Sum of Absolute Difference (SAD), Sum of Squared Difference (SSD), Zero Normalized Cross Correlation (ZNCC) and Normalized Cross Correlation (NCC) [17]. SAD was considered as the simplest cost function. In contrast, NCC has more computation complexity than SAD and SSD. [18] and [19] used different techniques to generate the depth map. [18] implement a dynamic programming-based method to estimate the vegetation height near power line towers.

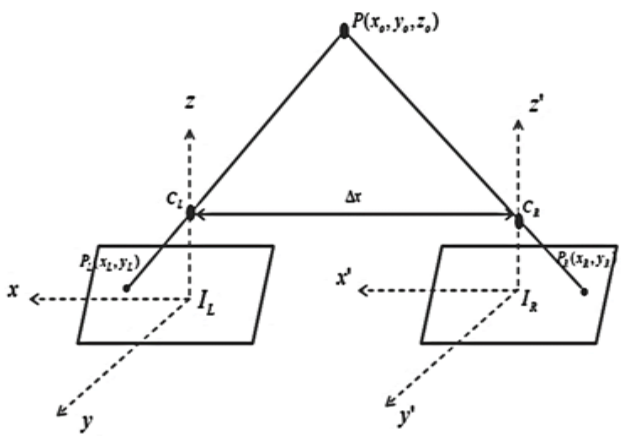

**Figure 4:** Stereo Image mode l [17]

Whereas [19] introduced a new graph-cut algorithm which can be used in satellite stereo images for creating the disparity map. The new Graph-cut algorithm can increase the accuracy of the disparity map by reducing the noise in regions that have no texture. The accuracy of the introduced algorithm reaches 80% compared to the existing graph-cut method.

### 3.4 Other Detection Method

Despite that most vegetation detection in power line corridor ROW used Vegetation Index (VI) and Stereo matching methods, several researchers have used different methods like change detection. [20] used detection algorithm based on probability and auto change where the algorithm classifies the satellite image into high and low vegetation region based on the image texture. The texture was scanned pixel by pixel while risk detection in the power line corridor was done by setting a threshold value, 7.5m for 20 KV and 26m for 110 KV. Therefore, any high vegetation activities within the danger zone will be captured. The vegetation height was estimated using laser scanning encroachment detection methods from satellite images. Figure 5 shows the Vegetation Encroachment Detection Methods in satellite imaged.

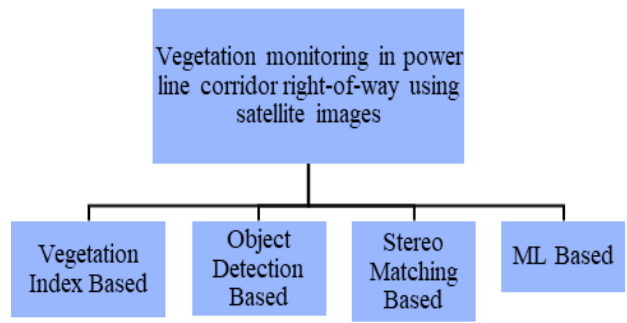

**Figure 5:** show the main vegetation encroachment detection methods that used in satellite image

### 4. MACHINE LEARNING METHODS

Machine Learning (ML) was considered as one of the AI branches, which can give the computer machines the ability to forecast and recognize patterns. ML basically uses three common methods: supervised methods that depends on pre-labeled data, unsupervised methods which has the ability to automatically categorize the data into classes based on the data patterns and reinforcement methods that can learn by observing the case environment. ML algorithms has been implemented in both satellite image and other types of remote sensing methods, In [21] they used non-satellite aerial images to train a Multi-Layer Perceptron (MLP) neuron network to classify four types of electrical power towers. Tanmay Prakash et all [22] used an Active Learning (AL) technique, which was based on supervised pre-labelled data to detect power transmission towers in satellite images. They extract the Haar-Like-Features and Local Binary Feature (LBF) as the main features to feed AdaBoost ML algorithm classifier. All the work was cloud-based with serval machines that worked simultaneously. The classifier produced 80% of accuracy and 50% for the recall accuracy. [23] introduced an image feature that is an extension from Higher-order Local Autocorrelation (HOAC) to extract both spatial and spectral relationship from multispectral satellite images. They also used AdaBoost as a classifier for the patch images and obtained a precision accuracy of approximately 90% and a recall accuracy less than 80%. ML algorithms were also implemented to detect vegetation cover in satellite images. [24] introduced a method to detect vegetation coverage by implementing both supervised and unsupervised ML algorithm by dividing the detection step into two parts: the supervised patch level classification step using Support Vector Regression (SVR) and unsupervised pixel-level classification. The method performs 98% accuracy. The study shows that ML can be very accurate for vegetation detection if appropriate classifier were used. Table shows the main difference between ML-based vegetation detection method compared with other methods

**Table 2:** Differences between ML methods and Current methods

| ML based methods | Current methods |
|---|---|
| Detection process depends on the dataset preparation and the classifier accuracy | Detection process depends on the threshold values. |
| Some ML algorithms need an extensive computation capability and hardware requirements like GPU, large memory and multithreading to perform the training process. | Usually, the algorithm can work under normal computation capabilities. |
| ML algorithms Can be very accurate if the training done in an effective way | The accuracy depends on the threshold values and the tuning parameters. |
| Creating and labeling datasets and training process can be very time consuming | No training process |
| Dynamic detection capabilities for multiple classes | Usually used for single object detection |

### 4.1 Deep Learning

Deep Learning (DL) is a branch of ML in which the Deep Neuron Networks (DNN) have more hidden layers than a regular Artificial Neuron Network (ANN). DL algorithms can be very accurate in object detection and pattern recognition [25]. In satellite image object detection, deep learning can be very effective to detect multiple objects in one image which can be a challenging task for normal methods. DL algorithms has been enhanced dramatically since the revolution in computer micro- processors and the graphic processor unit which can be used to reduce the time of training the DNN.
The Convolution Neural Network (CNN) represent a backbone structure in DL, Figure 6 shows the general CNN structure, where the input is the satellite images and the output is the class of the detected object. The hidden layers represent the feature map process were there many convolutional and pooling stages [26][27].

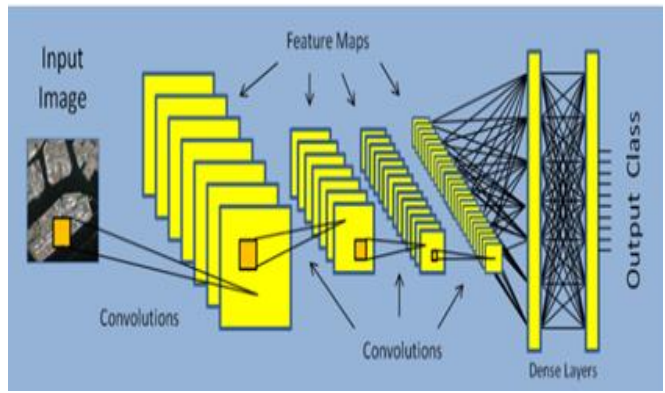

**Figure 6:** Shows CNN structure used for classification [26]

Many works have been done to detect transmission and distribution towers using UAV in [28] the authors used LS-Net fully CNN to detect transmission line from UAV image as well [29] used You Only Look Once (YOLO) DL algorithms to detect and localize transmission towers with different sizes in images taken by a cable walker drone. Also YOLO became a very popular method for object detection and localization in both normal and satellite images [30].
However, most of the works that have been implemented to detect power transmission towers used UAV machines like drones and fixed-wing autonomous planes [31],[32] and [33] Figure 7 Shows power transmission line detection and localization from UAV using DNN [34].

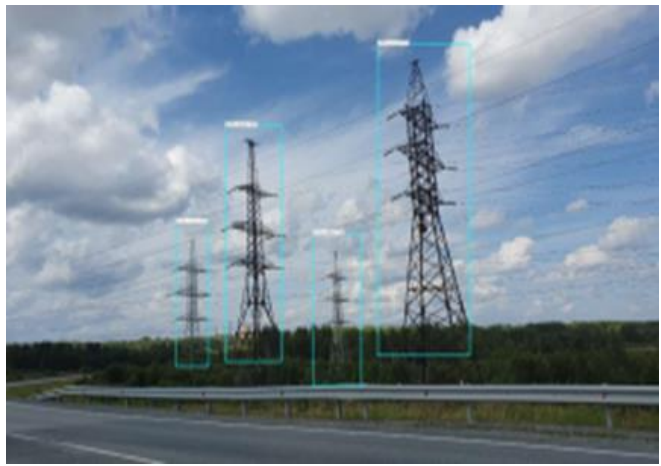

**Figure 7:** Shows power transmission line detection and localization from UAV using DNN [29]

DNN also has been used to detect and localize objects in satellite image. [35] introduced a study on modified U-net DL technique to detect multiple objects in satellite images. The algorithm was used to detect manmade structures like buildings, tracks, roads, trees, small and large vehicles, crops, Waterway and standing water. The classification done in pixel wise with testing accuracy of 0.981, and training accuracy of 0.9833. In [36] the researchers used an object detection methods called refined single-shot multi-box detector that based on deep learning to locate and detect object with different sizes in a panchromatic satellite images. Table 3 summarize the commonly used methods in vegetation detection and the types of satellite images used.

**Table 3:** Shows the commonly used detection methods and the type of satellite equipment

| Research | Methodology | Satellite type and resolution |
|---|---|---|
| Monitoring of Vegetation Near Power Lines Based on Dynamic Programming using Satellite Stereo Images [34] | Stereo matching Based (Dynamic programming) | QuickBird of 0.61m pan and 2.4 multispectral GSD |
| Monitoring powerline corridors with stereo satellite imagery [16] | Stereo matching Based (DEM)&(DSM) | QuickBird of 0.61m pan and 2.4 multispectral GSD |
| Vegetation Control of Transmission Lines Right-of-way [11] | VI based (ARVI) | QuickBird of 0.61m pan and 2.4 multispectral GSD |
| Study of Overhead Power Line Corridors on the Territory of Pskov Region (Russia) Based on Satellite Sounding Data [10] | VI based (NDVI) | Sentenile-2 with 10m multispectral GSD |
| Vegetation Coverage Detection from Very High Resolution Satellite Imagery [24] | Machine Learning Based (Supervised SVR patch wise) & (unsupervised pixel wise classification) | WorldView2 with 0.46 m Pan and 1.84m multispectral |
| Using a U-net convolutional neural network to map woody vegetation extent from high resolution satellite imagery across Queensland, Australia [37] | Machine Learning Based (Deep Learning) | Earth-imagery with 1m resolution |
| Designing of Disparity Map Based on Hierarchical Dynamic Programming Using Satellite Stereo Imagery [38] | Stereo matching Based (Hieratical Dynamic programming) | QuickBird of 0.61m pan and 2.4 multispectral GSD |
| Identification of Power Poles Based on Satellite Stereo Images Using Graph-Cut Algorithm [19] | Stereo Matching Based (modified graph-cut) | Pleiades with GSD of 0.7m for Panchromatic and 2.25m for multispectral |
| Active learning for designing detectors for infrequently occurring objects in wide-area satellite imagery [22] | Machine Learning Based (Supervised Active Learning) | WorldView with (0.4-0.5) m Pan and 2m multispectral |

## 5. CONCLUSION

Monitoring vegetation encroachment in power line corridor using satellite images can replace the other relatively expensive methods like LiDAR and airborne monitoring. There are many current techniques used for vegetation detection using satellite images like vegetation index-based methods, object detection-based methods and stereo matching-based methods. However, the current detection methods depend usually on manually adjusting threshold values which can not provide a dynamic way for detecting vegetation in different resolutions and different testing environments. On the other hand, Deep Learning can be very effective with high classification accuracy to detect multi objects in satellite images. which opens a promising solution in detecting vegetation encroachment in power line corridors.

## ACKNOWLEDGEMENT

This research is registered under Innovation and Research Management Center, University Tenaga National (UNITEN) under the BOLD 2020 grant scheme (Project Code: RJ010517844/051).